# Graph Transformer with Disease Subgraph Positional Encoding for Improved Comorbidity Prediction

Xihan Qin
Department of Computer and Information Sciences
*University of Delaware*
Newark, USA
xihan@udel.edu

Li Liao
Department of Computer and Information Sciences
*University of Delaware*
Newark, USA
liliao@udel.edu

*Abstract*—Comorbidity, the co-occurrence of multiple medical conditions in a single patient, profoundly impacts disease management and outcomes. Understanding these complex interconnections is crucial, especially in contexts where comorbidities exacerbate outcomes. Leveraging insights from the human interactome (HI) and advancements in graph-based methodologies, this study introduces Transformer with Subgraph Positional Encoding (TSPE) for disease comorbidity prediction. Inspired by Biologically Supervised Embedding (BSE), TSPE employs Transformer's attention mechanisms and Subgraph Positional Encoding (SPE) to capture interactions between nodes and disease associations. Our proposed SPE proves more effective than LPE, as used in Dwivedi et al.'s Graph Transformer, underscoring the importance of integrating clustering and disease-specific information for improved predictive accuracy. Evaluated on real clinical benchmark datasets (RR0 and RR1), TSPE demonstrates substantial performance enhancements over the state-of-the-art method, achieving up to 28.24% higher ROC AUC and 4.93% higher accuracy. This method shows promise for adaptation to other complex graph-based tasks and applications. The source code is available in the GitHub repository at: https://github.com/xihan-qin/TSPE-GraphTransformer.

## I. INTRODUCTION

Comorbidity, defined as the simultaneous presence of multiple medical conditions in a single patient [1], significantly influences disease management, therapeutic approaches, and prognostic outcomes [2]. For example, studies have shown that patients with specific comorbidities face various negative outcomes from COVID-19, highlighting the critical need for thorough comorbidity analysis in managing such pandemics [3-5]. Grasping the intricacies of comorbidity patterns is crucial for deciphering complex disease interconnections and uncovering common molecular pathways. The network-based methodologies [6-9] for predicting and examining these interrelationships significantly enhances our understanding and paves the way for innovative diagnostic and therapeutic strategies.

The human interactome (HI) [10] is a graph/network of collected protein-protein interactions in human cells. In the HI, nodes represent genes (or their protein products) and edges represent the interactions between proteins. Each node's connectivity and position within the network encapsulate important information relevant to the corresponding gene's role in the cellular processes. To extract such information, graph embedding techniques are employed to map nodes to a low-dimensional continuous vector space, preserving essential graph topology and making it compatible with machine learning models. This embedding representation enables more efficient and accurate analysis, facilitating downstream tasks such as node classification, link prediction, and, in our case, subgraph relationship study for disease comorbidity prediction.

Among the methods leveraging the HI to study disease comorbidity, a recent method, Biologically Supervised Embedding (BSE) [11], significantly outperforms the previous method, geodesic embedding (GE) [12], by using supervised selection to generate the most biologically relevant embedding vectors. The study of BSE highlights the critical role of disease associations and node connectivity in predicting comorbidities using the HI graph. Building upon these foundational concepts, this study explores the potential of the Transformer model to leverage node attention mechanisms and subgraph positional encoding for node connectivity and disease associations as emphasized by BSE for comorbidity prediction.

The Transformer model, introduced by Vaswani et al. in 2017 [13], has exerted a profound influence on the development of large language models. Its cornerstone is the multi-head attention mechanism, comprising both self-attention for the source language, self-attention for the target langue, and cross-attention for source-target language relationships. This architecture's ability to capture diverse aspects of relationships across different attention heads distinguishes it from recurrent neural networks, particularly due to its computational efficiency. Notably, renowned models like BERT [14] and GPT [15] have harnessed the Transformer's architecture to achieve state-of-the-art performance across various tasks. The Transformers' adeptness in capturing contextual information and long-range dependencies has rendered them a preferred choice not only in language processing but also in various domains, including image classification [16-18], medical data research tasks such as compound-protein interaction prediction [19, 20], and healthcare predictive analyses utilizing electronic health records [21]. Dwivedi et al. introduced a generalized form of the Transformer neural network architecture (GT) [22] applicable to arbitrary homogeneous graphs, expanding its utility beyond natural language processing (NLP) to graph-based tasks. This adaptation bridges the gap between the original Transformer, which is for sequential data like sentences in LLM, and graph neural networks, which operates via message-passing mechanisms. Notably, their work introduces local attentions among graph nodes and Laplacian eigenvectors as positional encodings in the proposed graph Transformer framework.

In this study, we introduced Transformer with Subgraph Positional Encoding (TSPE) for disease comorbidity prediction, inspired by insights from BSE [11]. TSPE leverages Transformer's attention mechanism to capture node interactions and integrates Subgraph Positional Encoding (SPE) method for disease association information. Node2Vec was utilized for generating node embeddings. Given the skewness in both benchmark datasets RR0 and RR1, we evaluated TSPE's performance using ROC AUC as the primary metric, which is well-suited for skewed datasets, and accuracy metrics as a secondary measure. TSPE demonstrated substantial improvements over the state-of-the-art BSE method with SVM classifier, achieving a 28.24% increase in ROC AUC and a 3.04% increase in accuracy for RR0, with average scores of 0.9489 for ROC AUC and 0.9069 for accuracy. For the RR1 benchmark dataset, TSPE showed a 15.40% increase in ROC AUC and a 4.93% increase in accuracy compared to the state-of-the-art, achieving scores of 0.8009 for ROC AUC and 0.7294 for accuracy.

## II. RELATED WORK

The Generalized Transformer Networks (GT) [22] propose two frameworks: one tailored for graphs without edge features, and the other designed for graphs with edge features. Both frameworks employ Laplacian Positional Encoding (LPE). In the language transformation task, each vector corresponds to a word token in a sentence. However, unlike the fixed order of words in a sentence resembling a linear graph, the nodes in a graph exhibit arbitrary ordering. Consequently, LPE is employed by directly adding it to the linearly transformed node features, to inject graph structural information into the Transformer framework. Dwivedi et al.'s ablation analysis reveals that LPE captures superior structural and positional information in comparison to the Weisfeiler Lehman-based absolute positional encoding (WL-PE) employed in GraphBERT [23].

The LPE process involves two key steps: initially, Laplacian normalization [24] of the graph is conducted, followed by Eigen Decomposition. Subsequently, the resulting eigenvectors are sorted based on their corresponding nonzero eigenvalues, with the k smallest eigenvectors chosen to serve as the positional encoding.

$$L = D - A \quad (1)$$

$$\tilde{L} = D^{-\frac{1}{2}} L D^{-\frac{1}{2}} \quad (2)$$

Equation 1 defines the Laplacian matrix, with D being the degree matrix and A the Adjacency matrix. Equation 2 is the normalized Laplacian (denoted as $\tilde{L}$) used by LPE.

$$\tilde{A} = D^{-\frac{1}{2}} A D^{-\frac{1}{2}} \quad (3)$$

Equation 3 is the definition of normalized Adjacency Matrix ($\tilde{A}$), where D is the degree matrix, A is the unnormalized Adjacency matrix.

$$\tilde{L} = I - \tilde{A} \quad (4)$$

By substituting equation 1 and 3 into equation 2, equation 4 is derived, which is used to calculate normalized Laplacian matrix ($\tilde{L}$).

$$\tilde{L} = U \Lambda U^T \quad (5)$$

Equation 5 represents the Eigen Decomposition of the normalized Laplacian matrix ($\tilde{L}$). In these equations, U is an n x n matrix comprising orthonormal eigenvectors ($u_1$, $u_2$, ......, $u_n$), and $\Lambda$ is an n x n diagonal matrix consisting of eigenvalues ($\lambda_1$, $\lambda_2$, ......, $\lambda_n$). Following this, the k eigenvectors corresponding to the k smallest eigenvalues are selected for LPE encoding, with the value of k determined by the user.

Another crucial aspect of the GT framework [22] is its consideration of graph sparsity. Given that real-world graphs are typically large and sparse, with numerous nodes but limited connectivity between them, Dwivedi et al. proposed focusing attention on locally connected nodes rather than performing full attention across all nodes, which is the attention mechanism used in language transformer models. Their results demonstrate that this localized attention mechanism surpasses two baseline models: GCN [25] and GAT [26]. While GAT employs attention on local neighborhood connections without considering the entire graph structure, GCN utilizes convolutional operations over the entire graph to aggregate information from neighboring nodes. Consequently, GT's focus on neighborhood attention, coupled with Laplacian positional encoding for graph structure information, bridges the gap between these two approaches, outperforming both.

TransformerGO [27], introduced by Leremie et al., applies the transformer framework to predict protein-protein interactions (PPIs) by capturing the semantic similarity between sets of Gene Ontology (GO) terms represented as a graph. Unlike the language transformer framework, TransformerGO employs unmasked Multi-Head Attention in the decoder, enabling the model to attend to subsequent positions in the GO term node sets for specific proteins. This attention mechanism is crucial for accurately predicting the relationships between proteins, distinguishing between positive and negative interactions across datasets from Saccharomyces cerevisiae and Homo sapiens. TransformerGO outperforms classic semantic similarity measures and state-of-the-art machine-learning-based approaches in PPI prediction.

Node2Vec is utilized in TransformerGO [27] to process graph data by transforming each protein's annotated GO terms into embedding vectors, which serve as input for TransformerGO. Node2Vec [31] is a widely used embedding method that leverages biased random walks to simulate both depth-first search (DFS) and breadth-first search (BFS) strategies. These random walks are guided by the Return Parameter (p) and the In-Out Parameter (q), which control the likelihood of revisiting nodes and exploring new neighborhoods, respectively. A higher value of p encourages the walk to remain close to the starting node, promoting DFS-like, local exploration. Conversely, a lower value of q favors exploration of nodes further from the starting point, promoting BFS-like, global exploration. The resulting random walks are then fed into a Skip-Gram model to produce embeddings, where nodes within a specified window size have closer embedding values.

## III. MATERIALS

The human interactome (HI) dataset, along with comorbid disease pairs and their clinically reported relative risk (RR)

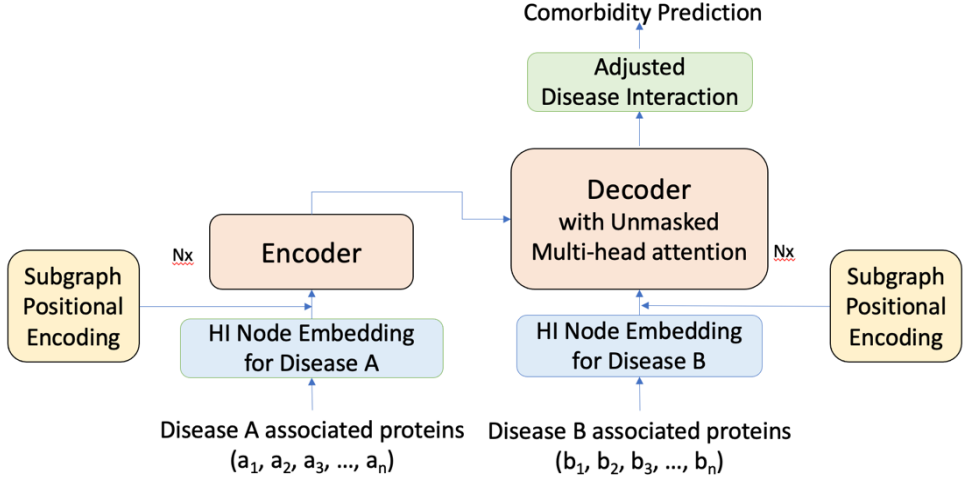

Fig. 1. Transformer with Subgraph Positional Encoding

values, were sourced from the study by Menche et al. [10]. These datasets are also utilized in the BSE study [11]. The RR score in [10] was calculated using a large patient medical history dataset, consisting of records from 13,039,018 patients diagnosed with one or more diseases over a 4-year period. The HI dataset, curated by domain experts, includes 13,460 protein nodes (denoted by gene IDs coding for proteins) and comprises 153 disease subgraphs. A total of 10,743 disease pairs were used in this study. These pairs were split into two different settings following the methodology of the BSE study [11] for pair comparison. In the first setting, disease pairs with an RR score greater than 0 were marked as positive for comorbidity, and those with an RR score less than or equal to 0 were marked as negative; this dataset is referred to as RR0. In the second setting, disease pairs with an RR score greater than 1 were marked as positive for comorbidity, and those with an RR score less than or equal to 1 were marked as negative; this dataset is referred to as RR1. In the RR0 dataset, 82.6% of the disease pairs are marked as positive, whereas in the RR1 dataset, 58.4% of the pairs are marked as positive.

## IV. METHODS

Drawing inspiration from the advancements made by BSE [11], Transformer with Subgraph Positional Encoding (TSPE) is proposed, with an architecture illustrated in Fig. 1. Our objective is to predict the relationship (whether comorbid or not) between two diseases, each is a subgraph in the HI, comprising a set of protein nodes associated with the corresponding disease. In this framework, understanding the connections between nodes within subgraphs is crucial for predicting disease subgraph relationships. Similar to how a language transformer model learns from the attention between words to comprehend sentences, the proposed framework learns from the attention between nodes to understand and predict relationships between subgraphs. However, unlike words in a sentence, nodes within subgraphs lack a fixed order. Therefore, we propose using subgraph positional encoding to inject each node's subgraph information, rather than a fixed order, into the framework.

As this task involves classifying the relationship between any two given disease subgraphs, rather than "translating" one subgraph into another, inspired by TransformerGO [27], the unmasked multi-head attention mechanism is employed in both the encoder and decoder. This ensures that the self-attention in decoder can learn to attend to all nodes within the subgraph, just as the encoder does. The output from the decoder maintains the same dimensions as the decoder's input. Each vector represents the interactions of proteins in disease B with every other protein in disease A, for a given disease pair (A and B). Since this is a binary classification task, the decoder's output is adjusted for binary classification, adopting the idea from TransformerCPI [28]. The equations below describe these processes.

$$s = \text{softmax}(\|X\|^2_{2,col}) \quad (6)$$

$$\|X\|^2_{2,col} = \left[\|X_{:,1}\|^2_2, \|X_{:,2}\|^2_2, \ldots, \|X_{:,n}\|^2_2\right] \quad (7)$$

Given an interaction matrix X, with dimensions m × n, which is the output from the decoder. A scoring vector s is first calculated using equation 6. The expression $\|X\|^2_{2,col}$ denotes the square of column-wise L2 norm, resulting in a vector of size 1× n, where n denotes the number of columns. Here, the L2 norm is calculated as the square root of the sum of the squares of its components applied to the column vectors. This can be expressed in equation 7, where $\|X_{:,j}\|^2_2$ represents the square of the L2 Norm of the j-th column. Subsequently, the softmax function is applied to normalize the values within the vector, ensuring that they sum up to 1. As a result, the scoring vector s has dimensions 1 × n.

$$\hat{y}' = \sum_{j=1}^{n}(Xdiag(s))_{.j} \quad (8)$$

As shown in equation 8, the scoring vector s is then used to weight each column in matrix X. Given s with elements $[s_1, s_2, \ldots, s_n]$ and size 1 × n, where each $s_i$ serves as a weight, and X $[X_1, X_2, \ldots, X_n]$ with size m x n, where each $X_i$ is a column of X, the operation $Xdiag(s)$ results in $[s_1X_1, s_2X_2, \ldots, s_nX_n]$ with a size of m x n. The candidate $\hat{y}'$ is then obtained by summing along n columns of the resulting matrix, yielding a vector of dimension m x 1. $(Xdiag(s))_{.j}$ represents the column-wise sum of the product of $Xdiag(s)$.

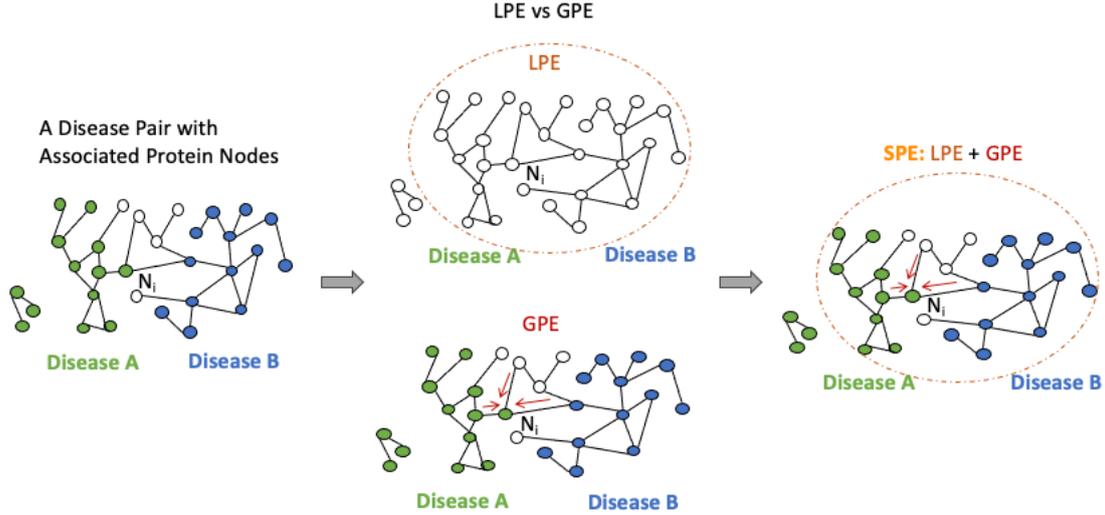

Fig. 2. Subgraph Positional Encoding Methods for Node $N_i$

LPE reaches out to all nodes in a single connected component, enclosed in the dashed circle, indiscriminate of the nodes' association with disease types. In contrast, GPE incorporates the nodes' disease associations, which are color-coded green and blue, and places greater emphasis on neighboring nodes, as indicated by the red arrows. SPE combines both LPE and GPE, not only captures further connections of the node but also incorporating the disease information gathered from the neighborhood.

$$\hat{y} = \sigma(W\hat{y}' + b) \quad (9)$$

$$L_{BCE}(\hat{y}, y) = -\frac{1}{N}\sum_{i=1}^{N}(y_i \log \hat{y} + (1 - y_i)\log(1 - \hat{y})) \quad (10)$$

The candidate $\hat{y}'$ is then passed through a linear layer followed by the sigmoid activation function to compute the prediction $\hat{y}$. Subsequently, the Binary Cross Entropy Loss (BCELoss), is employed to compare the result with the ground truth, yielding the loss for training. The Sigmoid layer and BCELoss are implemented together in PyTorch as a single class called BCEWithLogitsLoss [39].

In the HI dataset, neither node features nor edge features are available. Therefore, for TSPE, the input for each disease subgraph is obtained by first generating the graph embedding and then selecting the corresponding protein embedding vectors for each specific disease subgraph. Node2Vec [31] is employed to generate the node embeddings. We applied the Python Node2Vec package using their default setting with $p = 1$ and $q = 1$, so that the random walk behaves neutrally, resembling a standard random walk without a bias toward either local or global exploration. This balanced approach captures a mix of structural features but without a strong emphasis on either DFS-like or BFS-like behavior. And we set the window size to 2 to focus on capturing local relationships. The results obtained from TSPE are compared with those obtained using the SVM classifier in the state-of-the-art method.

The BSE study [11] underscored the significance of both disease association and node connectivity for comorbidity prediction. In our proposed framework, the node attention mechanism is used to capture node connectivity information. Given the absence of node features in the HI dataset, our node embeddings focus on representing the graph structure with an emphasis on local node connectivity. However, this approach lacks information about disease associations, specifically how gene nodes are categorized into different disease subgraphs.

To address this issue, we propose adding disease subgraph information as positional encoding, as shown in Fig. 2.

The Laplacian positional encoding (LPE) in the popular GT framework [22] selects eigenvectors associated with the smallest non-zero eigenvalues. These smallest eigenvalues are typically used for graph partitioning to identify clusters or communities within the graph, thereby incorporating cluster or subgraph information into the positional encoding. The authors of LPE attribute GT's success to this feature, as it outperformed the baseline models GCN [25] and GAT [26]. As illustrated in Fig. 2, while LPE can capture clustering information, it still lacks the ability to incorporate disease associations (i.e., the disease labels of each node). To address this limitation, we propose a novel approach, GPE, based on Graph Encoder Embedding (GEE) [32], to incorporate disease association.

GEE [32], proposed by Shen et al., shares similar properties with spectral embedding and is known for its rapid processing capabilities, capable of handling billions of edges within minutes on a standard PC [33]. GEE embedding Z is generated by a weight matrix, W, derived from known subgraph information and multiplied by either the Adjacency Matrix, Diagonal Augmented Adjacency Matrix, or Laplacian Normalization Matrix. The primary advantage of GEE lies in its utilization of the weight matrix W, which is directly based on subgraph information. providing more disease association information compared to Laplacian eigenvectors used in LPE [22]. Moreover, GEE is much faster to perform than the Laplacian Eigenvectors. The GEE embedding matrix, Z, constructed by the Adjacency Matrix, is defined as

$$Z = AW \quad (11)$$

Where A represents the Adjacency Matrix and W denotes the weight matrix. The size of the W matrix is N x K, with N representing the number of nodes and K indicating the number of label types. In our dataset, K is the total number of disease types, which is 153 diseases. For a node i belonging to disease subgraph j, $W_{ij}$ is calculated as $1/n_j$, where $n_j$ represents the number of nodes within disease subgraph j.

The challenge associated with using GEE stems from the size of K. There are two main reasons for this: firstly, if there are numerous subgraphs, K can become exceedingly large; secondly, when employing GEE-based positional encoding for summation with input node embedding, ensuring both embeddings possess the same dimensionality can be problematic, as they typically differ. To address this issue, we propose GEE Positional Encoding (GPE), which is outlined below.

$$Z = U\Sigma V^T \quad (12)$$

First, perform the Singular Value Decomposition (SVD) of Z. In Equation 12, U represents the left singular vectors, Σ is a diagonal matrix with singular values on the diagonal, and V represents the right singular vectors.

$$GPE = U_d \quad (13)$$

Then, as shown in Equation 13, select the first d largest left singular vectors based on the singular values to constitute the SPE embedding. The value of d is determined by the user.

Next, we propose Subgraph Positional Encoding (PSE) to integrate both the clustering information from LPE and the local disease label information from GPE. As depicted in Equation 14, for a node embedding matrix M (shown in the blue box in Fig. 1), GPE incorporates positional encoding by first adding LPE to M, and then concatenates with GPE to generate the final embedding E, which serves as the input to the Encoder and Decoder. This concatenation method enables flexible weighting of LPE and GPE contributions.

$$E = [(M + LPE),\ GPE] \quad (14)$$

TABLE I. TSPE PARAMETERS

| Layers | 3 |
|---|---|
| Learning Rate | 1e-04 |
| Batch Size | 20 |
| Valid | 0.1 |
| Dropout | 0.2 |
| Node Embedding Dimension | 64 |
| Number of heads | 8 |
| Position Encoding Dimension | GPE: 8 <br> LPE: 64 |

Overall, our TSPE model design (Fig. 1) utilizes the novel SPE (Fig.2) to encode key disease subgraph and clustering information from the graph into embedding representations. For each disease pair, the embeddings of nodes belonging to different diseases are input separately into the transformer framework's encoder and decoder. Through the attention mechanism, TSPE emphasizes key node connectivity, which is particularly valuable for revealing previously unknown connections absent in the HI data (currently estimated to be only 20% complete [10]) for the task. This approach highlights relationships within and between disease node sets, enabling effective comorbidity prediction on the test set.

We evaluate the performance of TSPE against the state-of-the-art BSE method [11] for comorbidity prediction. Given the skewness in both benchmark datasets, RR0 (82.6% positive comorbid pairs) and RR1 (58.4% positive pairs), we prioritize reliable metrics: ROC AUC as the primary metric and accuracy as the secondary metric. Additionally, we conduct an ablation analysis comparing TSPE to the same transformer framework without subgraph positional encoding or with the popular LPE only. Detailed hyperparameters and settings for TSPE used in the experiment are provided in Table 1.

V. RESULTS

In this section, all results are based on testing the RR0 and RR1 benchmark datasets using stratified 10-fold cross-validation. Each 10-fold split of the datasets underwent all designed experimental methods, and the performance metrics were averaged for comparison.

Table 2 compares TSPE against the state-of-the-art method [11], across different scenarios. Firstly, TSPE is compared with the SVM classifier used in [11], maintaining Node2Vec as the embedding method. Secondly, it is compared with BSE + Node2Vec + SVM. For the RR0 dataset, TSPE outperforms SVM significantly with a 41.80% increase in ROC_AUC and a 7.12% increase in accuracy. Compared to BSE with SVM, TSPE shows a 28.24% increase in ROC_AUC and a 3.04% increase in accuracy compared to the state-of-the-art, averaging 0.9489 for ROC AUC and 0.9069 for accuracy. Similarly, for the RR1 benchmark dataset, TSPE shows a 15.40% increase in ROC AUC and a 4.93% increase in accuracy compared to the state-of-the-art, reaching 0.8009 for ROC AUC and 0.7294 for accuracy.

Since the RR1 dataset exhibits lower metric scores compared to RR0, suggesting RR1's potential sensitivity to improvements with more accurate information in various positional encoding (PE) methods. We conducted an ablation analysis on PE to assess its impact on increasing metric scores using RR1 dataset. Table 3 presents the results of the TSPE framework with different PE methods on the RR1 benchmark dataset, measured using 10-fold cross-validation. Both PE methods show improvements in ROC AUC and Accuracy compared to the absence of PE.

The LPE method, introduced by Dwivedi [22], enhances ROC AUC by 0.36% and accuracy by 0.20% when integrated into our Transformer framework for 10-fold cross-validation. After removing one outlier, ROC AUC improves by 0.40% and accuracy by 0.20%. In Dwivedi's study [22], the focus was solely on accuracy for benchmark datasets (CLUSTER and PATTERN), reporting improvements of 1.03% and 0.859%, respectively. In our case, using LPE as positional encoding shows smaller improvements than those reported in [22]. This discrepancy may be due to LPE providing only clustering information in our task, lacking the disease subgraph label information, which is an important factor identified in BSE's work [11]. Our SPE method achieves the highest performance among all PE methods, with a 0.38% increase in ROC AUC and a 0.80% increase in accuracy for 10-fold cross-validation, and a 0.59% increase in ROC AUC and 0.80% increase in accuracy after removing one outlier. These results highlight that incorporating both clustering and local disease label information offers optimal positional encoding for our task, outperforming other methods.

TABLE II. METRIC SCORES COMPARED WITH BSE METHODS

| RR0 | | | |
|---|---|---|---|
| Metric | SVM | BSE_SVM | TSPE |
| ROC AUC | 0.5309 ± 0.0105 | 0.6665 ± 0.0301 | **0.9489 ± 0.0501** |
| Accuracy | 0.8357 ± 0.0039 | 0.8765 ± 0.0117 | **0.9069 ± 0.0683** |
| RR1 | | | |
| ROC AUC | 0.5497 ± 0.0079 | 0.6469 ± 0.0183 | **0.8009 ± 0.0152** |
| Accuracy | 0.6150 ± 0.0078 | 0.6801 ± 0.0166 | **0.7294 ± 0.0138** |

TABLE III. METRIC SCORES USING TSPE FRAMEWORK WITH DIFFERENT POSITIONAL ENCODING METHODS

| 10-Fold Cross-Validation | | | |
|---|---|---|---|
| Metric | NoPE | LPE | SPE |
| ROC AUC | 0.7971 ± 0.0146 | 0.8007 ± 0.0179 | **0.8009 ± 0.0152** |
| Accuracy | 0.7214 ± 0.0202 | 0.7234 ± 0.0202 | **0.7294 ± 0.0138** |
| After Removing One Outlier | | | |
| ROC AUC | 0.7951 ± 0.0171 | 0.7991 ± 0.0198 | **0.8010 ± 0.0218** |
| Accuracy | 0.7209 ± 0.0177 | 0.7229 ± 0.0230 | **0.7289 ± 0.0203** |

## VI. DISCUSSION

In this study, we introduced Transformer with Subgraph Positional Encoding (TSPE) for disease comorbidity prediction by leveraging the Transformer architecture, traditionally used in natural language processing tasks, and tailoring it for graph data and subgraph relationship classification. The innovation of TSPE, inspired by BSE's discoveries, is to utilize node attentions to capture key protein node connections and implement proposed subgraph positional encoding method to incorporate key disease association information for the task.

Our results on real clinical benchmark datasets demonstrate TSPE's significant superiority over the state-of-the-art BSE method with SVM classifier. By introducing GPE, we effectively reduce the dimensionality of GEE embeddings from 153 (the number of diseases in the dataset) to a customized dimension. Through our proposed SPE, which integrates both LPE and GPE, the model captures disease associations at the subgraph level. This integration with node embeddings allows for a comprehensive representation of both node connectivity and disease associations, enhancing the model's understanding of comorbidity relationships.

Among the positional encoding methods tested, SPE outperformed the popular LPE method, underscoring that integrating clustering and disease information provides the most effective positional encoding strategy for this task. This combined approach harnesses the strengths of both LPE and SPE, effectively capturing essential features for accurate comorbidity prediction.

Regarding limitations, if subgraph labels are unavailable, the proposed SPE cannot be applied. Users may still opt to use TSPE with "NoPE" option or with LPE or propose another positional encoding method to use with our transformer framework.

## VII. CONCLUSION

By integrating both protein node attentions and disease association information, TSPE achieves superior performance in comorbidity prediction compared to the state-of-the-art method using human interactome data. The ablation analysis on the positional encoding methods for TSPE demonstrated that SPE exhibits the highest efficacy, highlighting the value of combining clustering and disease information. Our findings suggest that the method of integrating positional encoding with node attentions, plays a critical role enhancing performance. Given its demonstrated success in predicting comorbidity with disease subgraphs, it is conceivable that TSPE can be adapted for other applications involving various subgraph relationship prediction tasks across different fields.

## ACKNOWLEDGMENT

We extend our gratitude to Dr. Joerg Menche for providing access to the comorbidity data. We also express appreciation to the anonymous reviewers for their invaluable feedback. The project was completed with computing resources partially supported by Sigma XI Grants in Aid of Research (GIAR).